\documentclass[journal]{IEEEtran}
\ifCLASSINFOpdf
\else
\fi
\makeatletter
\usepackage{hyperref}
\usepackage{multirow}
\usepackage{tablefootnote}
\usepackage{color}
\usepackage{epsfig}
\usepackage{epstopdf}
\usepackage{graphicx}
\usepackage{floatrow}
\usepackage{commath}
\usepackage{amsmath}
\usepackage{amsfonts}
\usepackage{float}
\usepackage[nolist]{acronym} % SB
\usepackage{mathptmx}
\usepackage[scaled=0.86]{helvet}

\newcommand*\titleheader[1]{\gdef\@titleheader{#1}}

\AtBeginDocument{%
  \let\st@red@title\@title
  \def\@title{%
    \bgroup\normalfont\large\centering\@titleheader\par\egroup
    \vskip1.5em\st@red@title}
}
\makeatother

\def\BibTeX{{\rm B\kern-.05em{\sc i\kern-.025em b}\kern-.08em
    T\kern-.1667em\lower.7ex\hbox{E}\kern-.125emX}}
\begin{acronym}
\acro{NN}{neural network}
\acro{RNN}{recurrent neural network}
\acro{GRU}{gated recurrent unit}
\acro{LSTM}{long short-term memory}
\acro{MFCC}{Mel-frequency cepstral coefficient}
\acro{PEM}{per-epoch noise mixing}
\acro{SNR}{signal-to-noise ratio}
\acro{ASR}{automatic speech recognition}
\acro{CNN}{convolutional neural network}
\acro{DNN}{deep neural network}
\acro{FNN}{feedforward neural network}
\acro{ACCAN}{accordion annealing}
\acro{WSJ}{Wall Street Journal}
\acro{CTC}{Connectionist Temporal Classification}
\acro{WFST}{Weighted Finite State Transducer}
\acro{WER}{word error rate}
\acro{LER}{character error rate}
\acro{LVCSR}{large-vocabulary continuous speech recognition}
\acro{ROI}{range of interest}
\acro{STAN}{sensor transformation attention network}
\acro{SER}{sequence error rate}
\acro{HMM}{Hidden Markov Model}
\acro{LER}{label error rate}
\acro{sMBR}{state-level minimum Bayes risk}
\acro{MLLT}{maximum likelihood linear transform}
\acro{fMLLR}{feature-space maximum likelihood linear regression}
\acro{MVDR}{minimum variance distortionless response}
\acro{STFT}{short-time Fourier transform}
\acro{ATTACC}{attention accuracy}
\acro{ATTCORR}{attention correlation}
\end{acronym}

\title{Data-Driven Neuromorphic DRAM-based CNN and RNN Accelerators}
\titleheader{Invited paper: 2019 IEEE Sig. Proc. Soc. Asilomar Conference on Signals, Systems, and Computers\\Nov. 3-6, 2019, Asilomar, CA, USA.
\href{https://www2.securecms.com/Asilomar2019/Papers/PublicSessionIndex3.asp?Sessionid=1003}{Session MP6b: Neuromorphic Computing (Invited)}}
\author{Tobi Delbruck and Shih-Chii Liu\\
Sensors Group, Institute of Neuroinformatics \\ University of Zurich and ETH Zurich, Switzerland\\
tobi@ini.uzh.ch, shih@ini.uzh.ch
}

\begin{document}

\newcommand{\mmsq}{\,mm$^2$}
\newcommand{\gops}{\,GOp/s}
\newcommand{\gopw}{\gops/W}
\newcommand{\tops}{\,TOp/s}
\newcommand{\topw}{\tops/W}
\newcommand{\gopa}{\,GOp/s/mm$^2$}

\maketitle

\begin{abstract}
The energy consumed by running large deep neural networks
(\textbf{DNNs}) on hardware accelerators
is dominated by the need for lots of fast memory to store both states and weights.
This large required memory is currently only economically viable through DRAM.
Although DRAM is high-throughput 
and low-cost memory (costing 20X less than SRAM), 
its long random access latency is bad for the unpredictable access
patterns in spiking neural networks (\textbf{SNNs}).
In addition, accessing data from DRAM costs orders of magnitude
more energy than doing arithmetic with that data. 
SNNs are energy-efficient if local memory is available and few spikes are generated. 
This paper reports on our developments over the last 5 years of
convolutional and recurrent deep neural network 
hardware accelerators that exploit either spatial or temporal 
sparsity similar to SNNs but achieve SOA throughput, 
power efficiency and latency even with the use of DRAM 
for the required storage of the weights and states of large DNNs. 
\end{abstract}

%\begin{IEEEkeywords}
%IEEE, IEEEtran, journal, \LaTeX, paper, template.
%\end{IEEEkeywords}

\IEEEpeerreviewmaketitle

There is currently interest from some mainstream 
AI communities in the relevance of neuromorphic engineering for artificial intelligence (AI).
In particular, can Spiking Neural Networks (SNNs) bring benefits to the Deep Neural Network (DNN) architectures used in AI?
%has been promoting the virtues of SNNs. 
These networks can be contrasted with conventional analog neural networks (\textbf{ANNs}) 
by their brain-inspired organization of interconnected spiking neurons to form DNNs. 
% In fact this is how we started our Neuromorphic Processor Project (NPP) that is funded by a big company. % with Samsung. 
Around 2015, we strongly believed that SNNs were key to making 
progress in addressing the huge inefficiencies in power consumption 
of hardware AI accelerators. %compared with biology.
We still believe this, but with a  more nuanced view of the
realities of what silicon and current memory technologies can offer. 

\begin{figure}[t]
    \centering
    \includegraphics[width=0.95\linewidth]{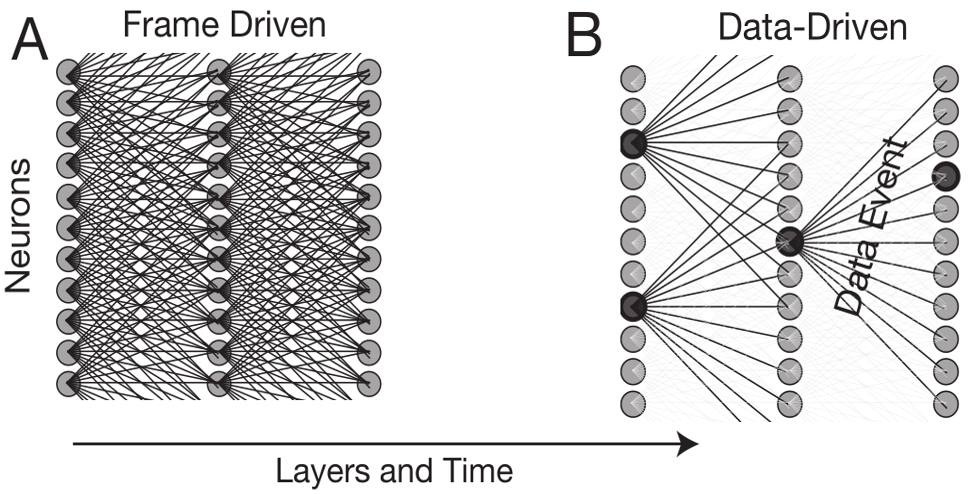}
    \caption{Concept of data-driven SNN-like approach for DNNs. 
    Instead of updating all neurons at each timestep (A), we only update a subset of neurons  (B).} 
  \label{fig:concept}
\end{figure}

The main contribution of this paper is to summarize 
for a lay audience our findings over the last 5 years of work 
towards the aim illustrated in Fig.~\ref{fig:concept}. 
We show that state-of-art (\textbf{SOA}) throughput, energy efficiency and latency 
can be achieved by convolutional and recurrent neural network (\textbf{CNN} and \textbf{RNN}) 
hardware accelerators that exploit data-driven synchronous architectures. We conclude by
comparing hardware SNNs with ANNs.

\textit{Note:} Our basic measure of 
computation (addition or multiplication) is the operation, or \textsf{Op}. 
When measuring DNN accelerator performance,
a multiply-accumulate (\textbf{MAC}) or synaptic accumulation operation is counted as 2\,\textsf{Op}.

\section{Sparsity in the brain's spiking neural networks}
\label{sec:sparsity}

\begin{figure}[hb]
    \centering
    \includegraphics[width=0.95\linewidth]{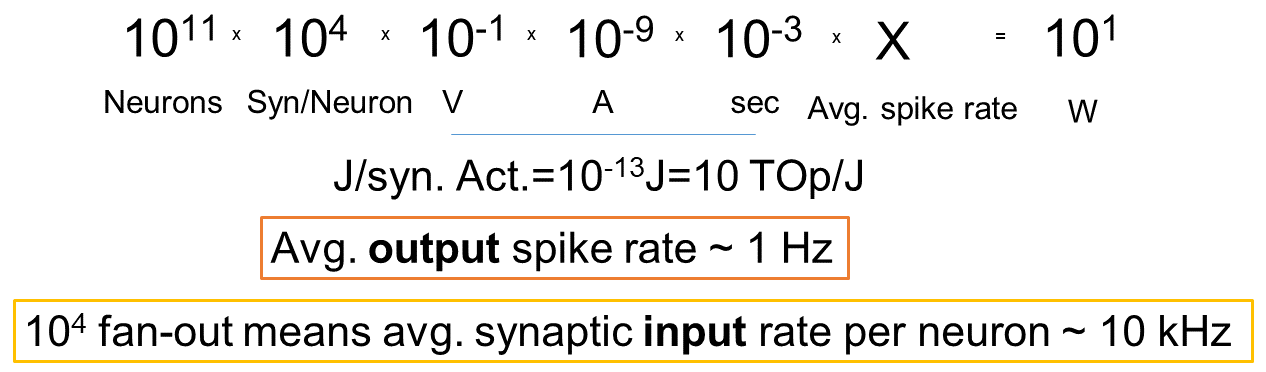}
    \caption{Sparsity and energy consumption of human brain.} 
  \label{fig:sparsity}
\end{figure}

Suppose we want to estimate the average 
spike rate $X$ of the brain's network of spiking neurons consuming $Y\sim 10 \rm W$ power. 
We will assume that every spike leads to one synaptic activation per synapse 
(i.e. we ignore significant synaptic activation failures of probably more than 50\%). 
If we multiply 
$X$ by the energy per synaptic activation, 
the number of synapses per neuron, and the number of neurons in the network, 
the result is the overall power consumption $Y$.
(Metabolism accounts for only about half the 
brain's energy consumption; about half is purely electrical~\cite{laughlin2003communication}.)
The first interesting results from this calculation (shown in 
Fig.~\ref{fig:sparsity}) is that the average spike rate $X$ is only about 1\,Hz. 
which is about 2 orders of magnitude lower than typical 
sensor sample rates. In other words, most DNNs that infer the meaning of sensor input
update every neuron much more frequently than 1\,Hz.

This calculation should not be taken too literally. 
Spike rates for active biological sensory input and motor output neurons are higher than 1\,Hz, 
and all the other numbers can
be off by a factor at least 2. 
The main point is that most 
brain energy is consumed by synaptic events, and only when they occur.

We find two more interesting observations from this napkin estimate: First, the 
large fan-out of $10^4$ 
means that neurons are still getting input at an average rate of 10\,kHz even with 1\,Hz average spike rate.
So they have the opportunity to spike with this much higher timing precision.
Second, the fan-out of $10^4$ is not so much higher than well-known CNNs: For example, consider
a CNN with 100 feature maps and 5x5 kernels. Then the fan-in to the next layer neurons will be 
2500, which is not much less than $10^4$.

Fig.~\ref{fig:sparsity} also calculates that the energy per biological synaptic 
update is about 100\,fJ, which 
equates to 10\topw. 
Recent SRAM-based low-precision 
DNN accelerators are approaching this 10\topw\ power efficiency,
but the match to the number computed in Fig.~\ref{fig:sparsity} is 
a bit misleading for two reasons: 
Firstly, one must assume that in the brain, 
each spike is precious and so it is likely 
that each synaptic operation contributes useful information to a computation, 
while in conventional DNNs, 
the operations are carried out no matter what their importance is. 
Secondly, a biological synaptic event
has complex dynamics 
that are certainly not present in a MAC.
The neurons in the ANN are equivalent to a point neuron which ignores the  
complex and nonlinear dendritic tree of a real neuron, and which almost certainly has 
rich computational properties that are barely appreciated. %except in some computational neuroscience circles.

\section{Our first hardware SNNs}
\label{sec:snn}

The fact that SNNs are inspired by the brain's spiking networks and arguments like the preceding suggest
significant advantages of SNNs compared with ANNs in that 
they access memory and compute only when needed.
Computer demonstrations 
showed advantages such as lower 
operation count and quicker output, e.g. as in~\cite{neil2016learning}. 

Our earliest developments of hardware SNN accelerators 
were in the CAVIAR project~\cite{CAVIAR2009}, a multi-chip, multi-partner project 
where one of the Spanish partners
implemented the first convolutional SNN ASIC~\cite{camunas2012event}. 
% This project also funded development of our first Dynamic Vision Sensor (DVS) 
% event camera~\cite{Lichtsteiner_etal_2008} which drove the convolution chips in the system.
CAVIAR drove the development of the first DVS event camera~\cite{Lichtsteiner_etal_2008},
and it was also a breakthrough by its demonstration 
of a fully-hardware mixed-signal convolutional SNN vision sensor system, but it would have been impractical 
for production because of the large number of ICs that it would need. The only way 
to increase the size of the network it could run would be to add more chips.

Our subsequent implementation of an SNN accelerator was called Minitaur~\cite{neil2014minitaur}. 
Minitaur was implemented on an FPGA and is our first event-driven digital implementation. 
It was  %hardware platform, 
%used DRAM for weight and state storage,
%Minitaur 
based on the premise that 
the network weights and states are stored in DRAM and local SRAM is used to cache 
the recently-used values, so that they can be quickly and cheaply accessed when needed.  
But Minitaur could not deliver a high throughput. 
On admittedly a low-performance 50\,MHz Spartan-3 FPGA, it could only handle
input spike rates of up to a few kHz when implementing 
a small 2-layer fully-connected network with a few thousand units and about 100,000 weights. 
The basic problem is that 
we cannot use SRAM alone to hold the parameters of a large network,
and to take advantage of sparsity, neurons should 
not need to fire repeatedly, but Minitaur relied on repeated
firing to have high cache hit rate.

% Subsequently, there have many great developments of hardware SNNs, including mixed signal
% ones like Neurogrid, ROLLs, and BrainDrop, and digital ones like TrueNorth and Loihi. We will discuss these later.

\section{SNNs can be accurate}
\label{sec:snnaccuracy}

Minitaur led to a very useful outcome in that it 
shaped our later work on building DNN accelerators.
It was the basis for an eventual successful proposal to an industry partner to explore SNNs and power-efficient ANNs.
At the time, the partner %was not particularly interested in SNNs; 
%and they 
strongly doubted if SNNs can achieve reasonable classification accuracy on a large dataset.
We therefore undertook a study to investigate if SNNs could produce equivalent accuracy on the same classification task as ANNs.
In this study, we first looked at a previously proposed method for converting a CNN using ReLUs to a spiking CNN~\cite{cao2014}, and then developed improved methods for converting  continuous-valued CNNs and multilayer perceptrons (MLPs) to equivalent-accurate SNNs~\cite{diehlneil2015fast}.  
We later showed that we can also train for low-latency, low-compute SNNs~\cite{neil2016learning}.
%After our early experiments on the conversion, we realized that a simpler method that takes advantage
%of supervised training of ANNs would be to simply convert pre-trained ANNs to equivalent SNNs.
%Following a previous study that showed possible conversion to spiking CNNs~\cite{cao2014}, 
Our student Bodo Rueckauer took this on, and
 developed new conversion methods that improve on the classification accuracy of deeper SNNs.
 %a set of handy conversion tools that can 
These methods show that larger pre-trained ANNs (e.g. VGG-16 and GoogLeNet Inception-V3) can be converted into equivalent SNNs that have the best SOA accuracy results at the time~\cite{rueckauer2017theory}.
%Using this ANN-SNN conversion tool, we showed that even deep CNNs like GoogleNet can achieve almost the same accuracy as the original CNN.

 We also demonstrated that SNNs could be trained 
using backpropagation to SOA classification accuracy on MNIST~\cite{lee2016training}.
Researchers are now exploring many interesting ideas
for training SNNs using local learning rules, or approximations using spike rates during backprop.
\section{Two problems with SNNs}
\label{sec:badthingsaboutsnns}

It turned out that there were two big problems with these results in relation to silicon implementation of SNNs:
\begin{enumerate}
    \item Although we can convert pre-trained ANNs %trained by supervised learning 
    to equivalent SNNs to achieve almost equal accuracy as ANNs, 
     the conversion equates activations to spike rates. 
    The resulting SNN spike rates requires
    sending many spikes per activation value 
    when the activation value is high. By contrast, a standard ANN sends just the activation value itself.
    The result was that the SNN was usually \textit{more} costly to run than the ANN, 
    \label{sec:lotsofspikes}
    i.e. required more synaptic operations than the MACs in the ANN.
    \item As we will discuss in Sec.~\ref{sec:snnanncomparison}, the resulting asynchronous unpredictable spikes made memory access 
    unpredictable, which ruled out using DRAM memory, but only DRAM is affordable for scaling the hardware
    to big DNNs (SRAM is 20X more area than DRAM).
    \label{sec:asynchmemaccess}
\end{enumerate}

Since our observations of point \ref{sec:lotsofspikes}, tools like~\cite{Shrestha2018-yj}
have enabled more people to train SNNs using backprop and were used to show that
at least for shallow classifiers---where 
low precision activations provide sufficient 
accuracy---SNNs can achieve equivalent accuracy as ANNs with fewer synaptic operations.
But Minitaur and point~\ref{sec:asynchmemaccess} stuck in our minds. We asked,
would it possible to make use of sparsity in a synchronous DNN? 
Before going into these developments, we make a slight digression to review the architecture of DRAM.

\section{DRAM access must be predictable}
\label{sec:dram}

% The economics of memory types and the 
% fact that DRAM needs to be read in a structured manner 
% for speed and best energy efficiency is well known to electrical and computer engineers, 
% but we point it out here again because it is seems to be ignored in the neuromorphic community.

In dynamic random access memory, a bit is stored using a single transistor switch as charge on a capacitor. 
Since the charge leaks away, it must be periodically refreshed by reading and restoring the bit.
DRAM architecture has two important features, one good and one bad.

Firstly, because a DRAM bit cell needs only a single transistor and a highly optimized cylindrical capacitor, 
it is very tiny and therefore cheap.
By contrast, a static random access memory (SRAM) cell 
is constructed as a bistable latch circuit from a minimum of 6 transistors. 
It makes an SRAM cell area about 20X larger than a DRAM cell. 
Although the digital nature of SRAM state makes it very fast to
write and read, it is costly compared with DRAM for storing large amounts of data.

\begin{figure}[tb]
    \centering
    \includegraphics[width=0.95\linewidth]{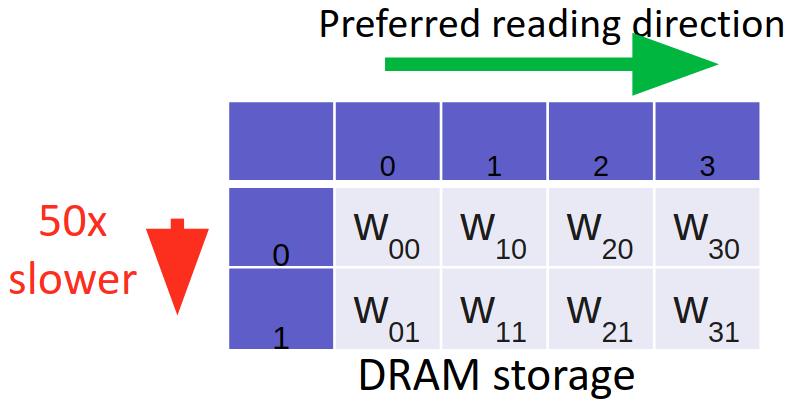}
    \caption{DRAM memory can be read in burst mode very quickly,
    but changing rows takes 50 times longer for DDR3 DRAM. 
    (Estimate courtesy of X. Chen from Vivado simulations of Zynq MMP)} 
  \label{fig:dram}
\end{figure}

Secondly, reading the bit of charge stored in a DRAM cell takes time, 
because it is done by putting a special type of metastable latch called a sense amplifier 
into its metastable point and then dumping the charge onto the latch's input. 
Since the latch is large compared to a bit cell, and the sensing takes time, 
sensing the charge is done in parallel at the bottom of long columns of hundreds of DRAM bit cells. 
To read DRAM memory, you address with the `word line' a row of cells.  The sense amps 
connected to the column `bit lines' then read the bits stored in the cells of that row. 
The sense amplifier digital outputs are then read out very fast using burst mode. 
If the DRAM reads and writes can be scheduled, then DRAM achieves high bandwidth.
But reading DRAM memory randomly is much slower (e.g. 50X slower in the case of DDR3 DRAM)
than reading it out row by row, at least in terms of throughput. 
This extreme asymmetry is illustrated in Fig.~\ref{fig:dram}. To summarize, accessing DRAM is only fast if the 
IO can be scheduled to consist of series of predictable pipelined burst transfers.

\section{Accelerating CNNs by exploiting spatial activation sparsity}
\label{sec:cnn}

Since we wanted an accelerator that could scale itself to 
any sized network, we needed to use DRAM, but this ruled out
`pure' asynchronous SNNs 
because of their unpredictable memory access.
But we thought that it might still be possible 
to exploit activation sparsity.

% ,
% based the observation 
% that many zeros appear in the feature maps after applying the
% Rectified Linear Unit (\textbf{ReLU}) activation function on the output of the neurons.  
% The first use of ReLU in an ANN was based on observations of measured 
% threshold-linear firing rate curves in neuroscience experiments~\cite{hahnloser2000digital} 
% and a decade later was adopted (without realization of this previous work)
% by the mainstream machine learning 
% community to replace previous sigmoid units~\cite{Jarrett2009-fr,nair2010rectified}; 
% and is now ubiquitous in CNNs.
\begin{figure}[tb]
    \centering
    \includegraphics[width=0.9\linewidth]{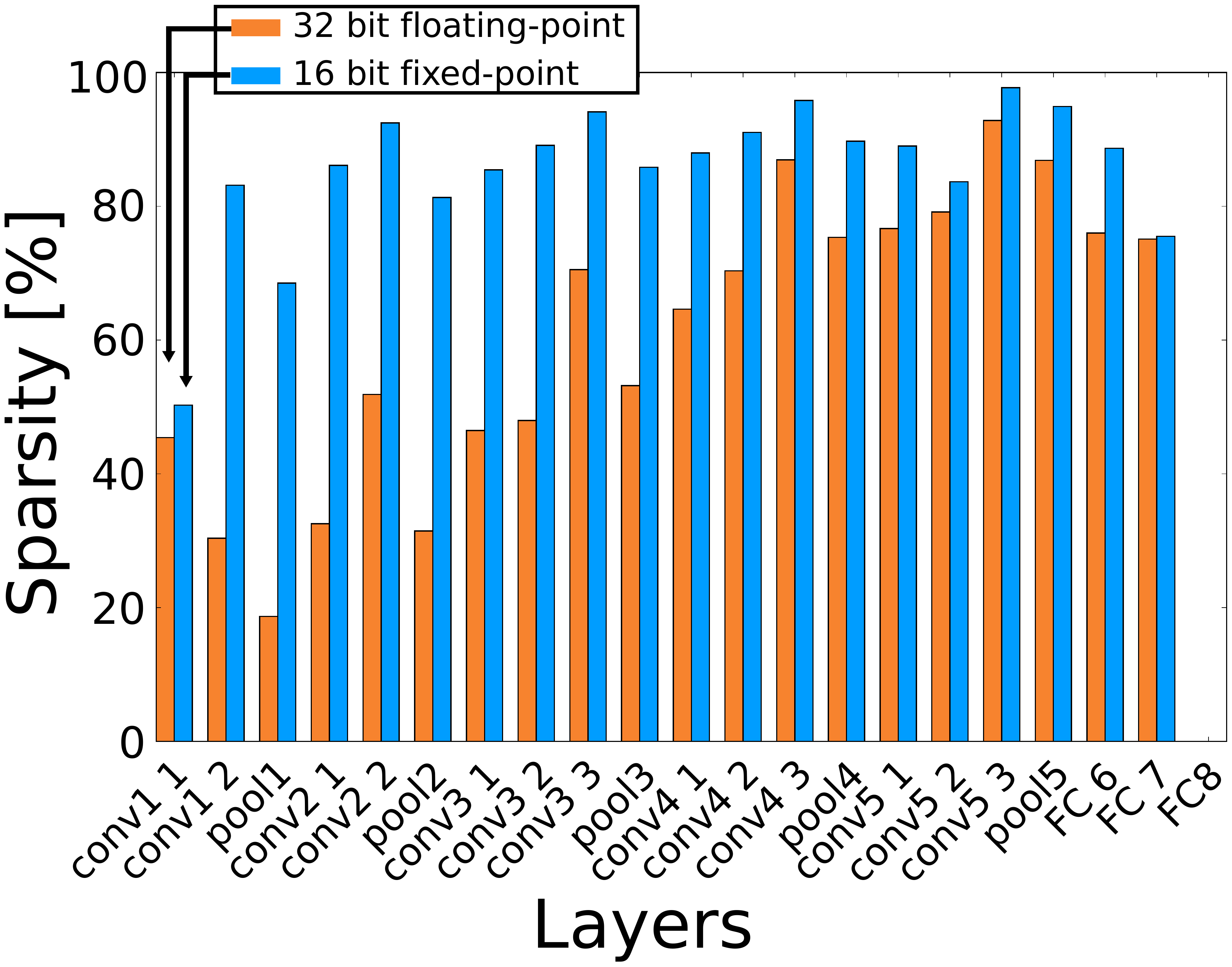}
    \caption{Activation sparsity of VGG-16 CNN image classifier layers 
    when trained to floating-point or fixed-point precision, 
    averaged over 1000 random images. 
    Adapted from~\cite{aimar_nullhop:_2018}}. 
  \label{fig:relusparsity}
\end{figure}

Although the Rectified Linear Unit (\textbf{ReLU}) activation function
sprang from
neuroscience~\cite{hahnloser2000digital}, it was not until about
2010~\cite{Jarrett2009-fr,nair2010rectified} that the 
ReLU has widely been used in CNNs.
Since ReLU output is zero anytime the input is non-positive,
using it results in
many zeros in feature maps. These zero pixels 
are like SNN neurons that are 
not spiking. They cannot have 
any downstream influence, so why bother
to compute their MACs?  But 
how many zeros are there in practice?

Fig.~\ref{fig:relusparsity} show measurements of
activation sparsity across layers in a classifier CNN. 
If the network is trained and run with floating point precision, 
the sparsity is about 50\%~\cite{milde2017adaption}. While developing tools
for quantizing weights, we discovered that
if the network is trained
to 16-bit weight and state precision
by using a quantization technique
called dual-copy 
rounding or Pow-2 ternarizaton~\cite{stromatiasneil2015robustness,ott2016RNN},
the average sparsity rises to nearly 80\%.
It means that 4/5 of the pixels are zero.
It seems that
the quantization training
increases activation sparsity.
Whatever the reason, in principle, we can skip over all
these MACs and easily obtain a speedup of a factor of 4.

The first published CNN accelerator to exploit activation 
sparsity was the well-known \textit{EyeRiss}~\cite{Chen2017-ss}.
Its dataflow architecture
used each DRAM access for hundreds of MACs, but it
did not take full advantage of ReLU sparsity 
because it decompressed and recompressed 
the run-length-encoded feature maps and just power-gated the MAC units.

\begin{figure}[tb]
    \centering
    \includegraphics[width=0.92\linewidth]{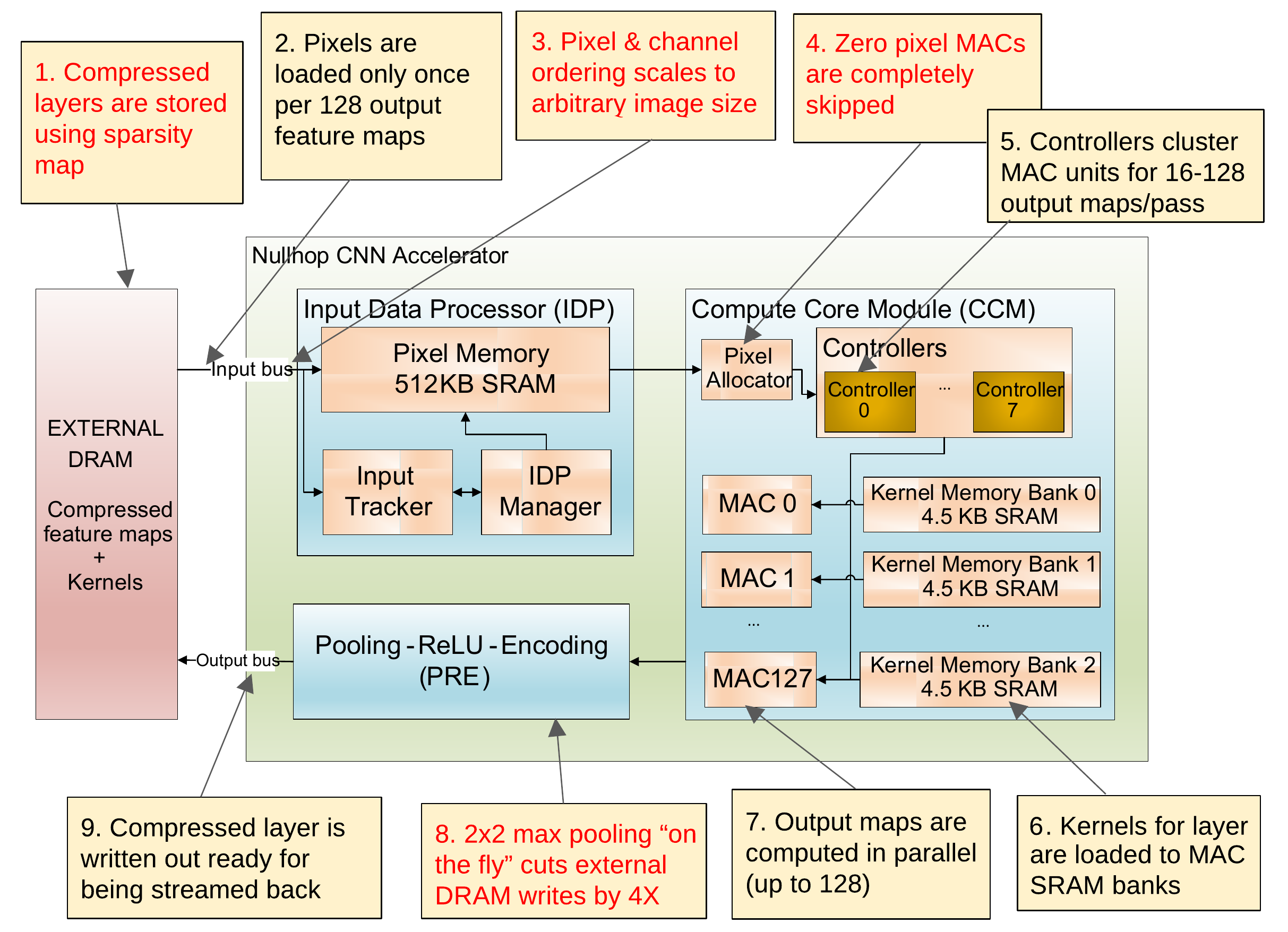}
    \caption{Architecture of NullHop CNN accelerator.} 
  \label{fig:nullhoparch}
\end{figure}

Development of our \textit{NullHop} CNN accelerator was started by postdoc H. Mostafa, 
and then taken over by our student Alessandro Aimar~\cite{aimar_nullhop:_2018}. 
Nullhop is more efficient than Eyeriss by using 
several features illustrated in Fig.~\ref{fig:nullhoparch}.
It stores the feature maps using a sparsity map (\textbf{SM}) 
compression scheme.  
The SM is a bitmap that marks non-zero feature map pixels 
with a '1' bit, and stores the non-zero values in a 
non zero value list (\textbf{NZVL}).
That way it uses only 1 bit for inactive pixels and the full precision of 
16 bits in the NZVL, plus 1 SM bit for active pixels. 
The logic that processes the input feature maps to generate output feature maps 
never decompresses the feature maps 
and skips over all zero pixels without using any clock cycles.
Its pooling-ReLU unit saves a lot of intermediate memory access when max-pooling
feature maps.

Since the 
DRAM DDR3/4 memory interface IP block  is too expensive for us,
our reported NullHop results are from 
a 28nm process technology simulation and the System-on-Chip FPGA implementation.
In 28nm technology, NullHop would achieve a core power efficiency 
of about 3\topw\  with area efficiency of about 100\gopa. Factoring in DDR3 DRAM, the system
level efficiency would be about 1.5\topw\  with 
area of about 6\mmsq\  and throughput of about 500\gops.  
By 
exploiting the spatial activation sparsity, 
NullHop achieves an efficiency of almost 400\% for large image classification networks, 
by skipping MACs for zero activations. 400\% efficiency means that each MAC unit does the work of 4 MAC units
that do not exploit sparsity.

The efficiency is worth comparing to the brain sparsity estimate from above. 
In contrast to the brain power efficiency estimate, 
where we only count actual synaptic operations, for accelerators that exploit sparsity,
an operation is counted even 
if it is skipped. 
% because sparsity is exploiting to skip them. 
It is a fair 
comparison with accelerators that 
do not exploit data sparsity.
% The sparsity estimates for brain energy efficiency of about 10\topw\  from Sec.~\ref{sec:sparsity} are thus not comparable. 
%Since NullHop, Aimar has continued to develop CNN accelerators that exploit activation sparsity. 

\begin{figure}[tb]
    \centering
    \includegraphics[width=1\textwidth]{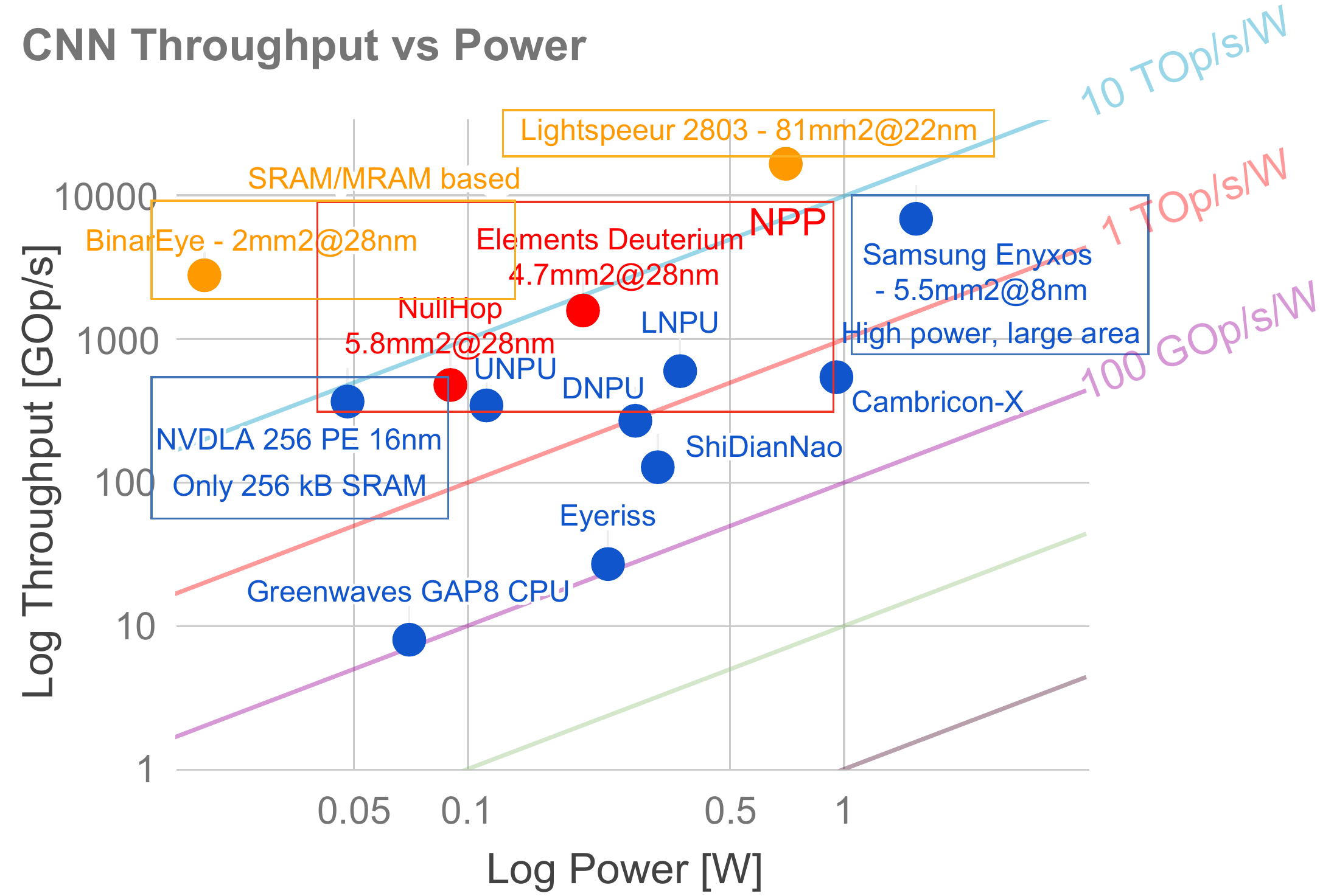}
    % https://docs.google.com/spreadsheets/d/1aA4mAvUrv7mbYH4zycg7eQrdvd8SkA5doPCGeCJr_eU/edit#gid=275363585
    \caption{Throughput versus power consumption for CNN ASIC accelerators: 
    Eyeriss~\cite{Chen_etal16}; 
    BinarEye~\cite{binarEye2018}; 
    DNPU~\cite{ShinDNPU2017}; 
    LNPU~\cite{YooLNPU2019};
    UNPU~\cite{LeeUNPU2018};
    Cambricon-X~\cite{zhangcambriconx2016};
    ShiDianNao~\cite{ShiDianNao2015}.
    (Figure generated by A. Aimar.)} 
  \label{fig:cnnperf}
\end{figure}

Fig.~\ref{fig:cnnperf} compares CNN throughput versus power for a number 
of published and unpublished accelerators. The CNN accelerators that Aimar is currently 
developing, called \textit{Elements},  
include both kernel compression and flexible bit precision.
The power reported in
this chart does not include DRAM power for DRAM-based accelerators.
The \textit{NullHop} and \textit{Elements} accelerators are labelled \textsf{NPP}.
The ones labeled ``\textsf{SRAM/MRAM based}'' use only on-chip memory. 
They achieve high power efficiency but are either very limited
(\textit{BinarEye}~\cite{binarEye2018}) or expensive (\textit{Lightspear}). 
Aimar's \textit{Elements} modeling shows that including DRAM limits power efficiency to around 2\topw\ 
(based on DDR3 energies), and publications hint to similar limits. 
So, despite the impressive amortization of DRAM memory power by 
sharing the weights and activations across all the involved operations, CNN accelerators are hitting
a memory bottleneck that is ultimately limited by DRAM energy itself.

\section{Accelerating RNNs by exploiting temporal activation sparsity in Delta Nets}
\label{sec:deltanets}

Nullhop was a step in the direction of an SNN-like synchronous accelerator, but 
it did not exploit any temporal sparsity. Was it possible to use this idea for CNNs, by making
them stateful and only updating the units that change?
Despite the publication of~\cite{oconnor2016sigmadelta}, we quickly discovered that in most cases it was not beneficial. %, because %it would mean doubling the DRAM bandwidth. 
Because a CNN is feedforward and stateless, it turns out that it is nearly always cheaper 
to completely compute the output of a layer by using the output of the previous layer,
rather than propagating changes.
%just reconstruct the network layer by layer from its input.
To compute state changes, one would need to %Instead of just reading the previous layer's activation to produce the next layer's activations, now we would need to 
store each layer's activations, and then read them before (possibly) updating them.
It means that we would need to store, read, and write large parts of the 
entire network state rather than recreating it 
for each new input frame. 

Stateless CNNs only benefit from exploiting temporal sparsity when this sparsity is 
very high, e.g. as in some surveillance applications~\cite{cavigelli2017cbinfer}, but
the architectures that benefit hugely from exploiting temporal sparsity are 
RNNs. In contrast with CNNs, RNNs are stateful and the state space is 
rather small, since it generally consists of vectors representing e.g., in the input case, 
an audio spectrogram window or a set of low dimensional sensor signals. 
In contrast with 
CNNs, 
where each weight can be used many times, 
the big weight matrices in RNNs that connect the layers take a lot of 
memory bandwidth to read, and each weight is only used once per
update. 
By using an RNN batch size $>1$, 
or by processing multiple RNNs in parallel, it is possible to reuse weights,
but we are mainly interested in real-time applications 
on embedded platforms 
where we get a new input sample and need to compute the RNN response to that sample as
quickly as possible.

\begin{figure}[tb]
    \centering
    \includegraphics[width=0.9\textwidth]{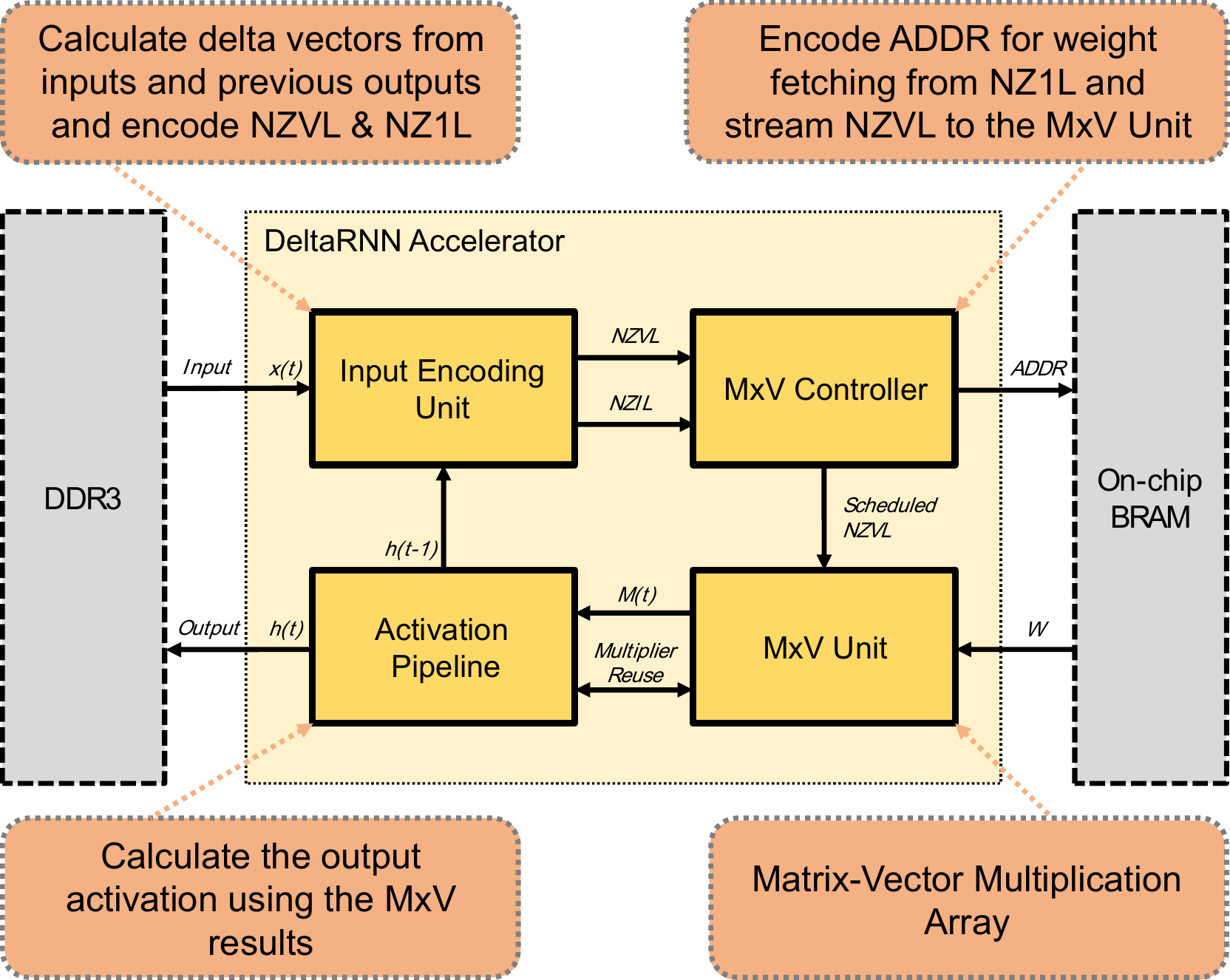}
    \caption{Block diagram of DeltaRNN accelerator on FPGA. MxV=Matrix-Vector Multiplication; \textbf{NZVL}=Non-Zero Value List; \textbf{NZ1L}=Non-Zero "1" List} 
  \label{fig:drnnblock}
\end{figure}

Thus we asked ourselves if it would be possible to propagate only \textit{significant 
changes} through an RNN without losing accuracy. 
The answer came from our student Daniel Neil in our
 DeltaNet paper~\cite{neil2016delta}. He showed that 
you can get away with only
updating units downstream from 
those whose activity changes by more than \textit{delta threshold} $\theta$ without
huge accuracy loss. 
And if you include  
$\theta$ in the forward pass during training, it helps the RNN to maintain higher prediction accuracy.
The $\theta$ is bit like a spike threshold, but it is evaluated
synchronously in the DeltaNet, and the DeltaNet
sends the full analog value of the activation to downstream units.
Using these delta events requires storing the pre-activation states of the neurons
rather than the usual post activation values, but the cost in memory or operations is not increased.

Using this DeltaNet principle
reduces memory access by 5X-100X, depending on the input statistics.
Thus a DeltaRNN sparsely computes in the temporal domain
like a spiking neural network (SNN), but “plays nice” with DRAM by allowing a more
predictable synchronous memory access. 

Our student Chang Gao has implemented several generations of DeltaRNN on FPGAs. 
Fig.~\ref{fig:drnnblock} shows a high level description 
of the blocks for the first published implementation~\cite{GaoDeltaRNN2018}. 
It achieved a throughput of 1.2\tops\ on 
wall plug power of about 15\,W.
This first implementation used FPGA block 
RAM (BRAM) rather than DRAM for the weights. His subsequent implementation for
portable edge computing which we call \textit{EdgeDRNN}~\cite{gaoedgedrnn2019} 
runs on a \$89 FPGA board. 
It uses DRAM
for the weights.
Running a 2L-768H-DeltaGRU, it
achieves a  mean effective  throughput of 20.2\gops\  and a wall plug 
power efficiency of about 8\gopw\, which is higher by at least a factor of at least four compared
to the NVIDIA Jetson Nano,  Jetson TX2 and  
Intel Neural Compute Stick 2, for a batch size of 1.
It is also at least six times quicker than these devices.
Since more than 80\% of EdgeDRNN power is for things like the ARM core 
processors, UART, and other support, the overall power efficiency is 
on the lower end of published RNN accelerators, but it remarkably 
achieves with 2.4\,W wall plug power the same latency as a 92\,W Nvidia GTX\,1080 GPU.

\begin{figure}[tb]
    \centering
    \includegraphics[width=\textwidth]{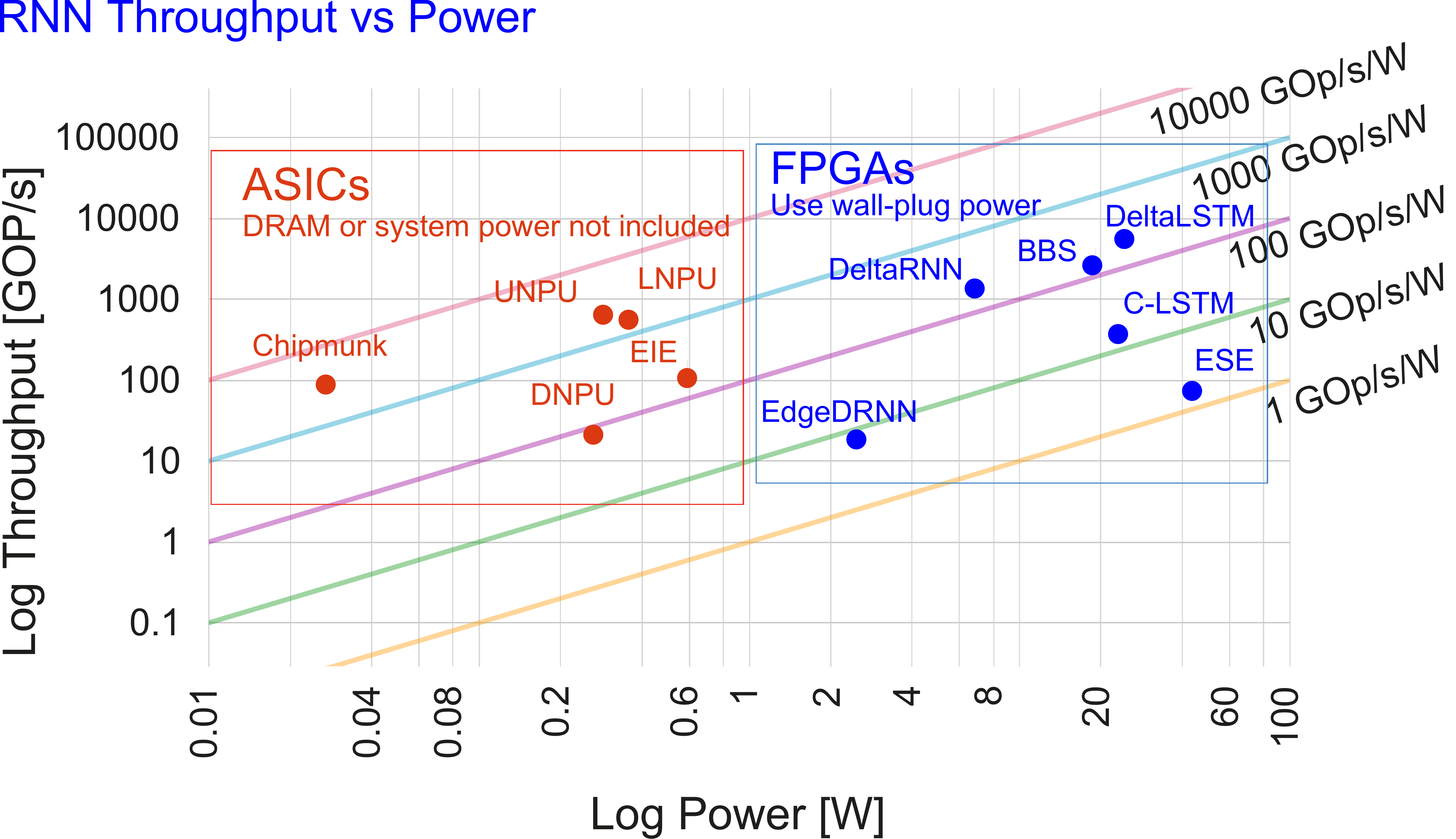}
    \caption{Throughput versus power consumption for FPGA and ASIC RNN accelerators. 
    ASIC implementations do not include DRAM memory power. 
    FPGA implementations are based on wall-plug power. 
    DNPU~\cite{ShinDNPU2017}; LNPU~\cite{YooLNPU2019}; 
    UNPU~\cite{LeeUNPU2018}; 
    Chipmunk~\cite{conti2018chipmunk}; 
    ESE~\cite{han2016ese}, 
    DeltaRNN~\cite{GaoDeltaRNN2018}; 
    C-LSTM~\cite{wang2018clstm};
    EIE~\cite{han2016eie};
    BSS~\cite{cao2019efficient};
    The DeltaLSTM variant 
    and the 
    EdgeDRNN~\cite{gaoedgedrnn2019} are being developed in our group.
    (Figure generated by C. Gao.)} 
  \label{fig:rnnperf}
\end{figure}

Fig.~\ref{fig:rnnperf} compares RNN throughput versus power 
for a number of published and unpublished accelerators.
The ASIC publications omit DRAM and system power 
so they would be pushed way over to the right if these were included.
Like for CNNs, RNN accelerators seem to be hitting an
efficiency wall of a few hundred \gopw\ that is determined by DRAM energy.
It seems that MRAM-based RNN accelerators are an attractive target if
fixed memory size is OK.

\section{Features and FOMs}
\label{sec:snnanncomparison}

Certain features make DRAM-based accelerators attractive, especially for system integration:
 \begin{itemize}
    \item Flexibility in DNN size, so it can run a range of networks, without much change in power efficiency, and so that
    the cost in memory is only enough to run the particular network.  
    \item Flexibility in DNN type, so that can 
    handle a tasks ranging from classification, detection, semantic pixel labeling, and video processing, requiring
    various levels of weight and especially state precision.
    \item Cost sharing in a system environment, e.g. ability to share existing  memory and (and expensive) memory interface.
\end{itemize}

An accelerator can be measured by a combination of figures-of-merit (\textbf{FOM}s) that rank how well it achieves
certain objectives:

\begin{enumerate}
    \item 
    it must be cheap, i.e. have 
high Op/s/mm$^2$ area efficiency,
    \item it must be fast enough, i.e. achieve sufficient Op/s throughput,
   \item it must be cool enough at this throughput, i.e. achieve high Op/s/W power efficiency.
\end{enumerate}

Synchronous CNN accelerators achieve throughput of 
about 1\tops\ with area around 10\mmsq, i.e. area efficiency of 100\gopa.
High efficiency is achieved by a high clock frequency and keeping the multipliers full of data.
% SNNs are much worse because they use most of 
% their area to hold weights and (mostly unchanging) for states.

% They come with a wide range of throughput, ranging from a few \gops\  to tens of \tops, depending
% on the memory bandwidth and number of processing elements.

Current DRAM-based CNN accelerator power 
efficiency seems to be asymptotically converging around few \topw,
and more memory bound RNN accelerators are hitting about 100\gopw. 
% There seems to be some upper bound that is 
% mainly determined by the latest DRAM memory technology.
% Surprisingly, it seems be quite comparable to SNNs: 
% SOA mixed-signal~\cite{Moradi2018-zw} and digital~\cite{Davies2018-cx,Cassidy13} SNNs  
% burn about 10\,pJ per synaptic activation; i.e. they 
% get about
% 100\gopw\  power efficiency.
% If we account for sparsity by generously multiplying
% by a factor of 10, it comes to about 1\topw.

DRAM-based CNN accelerators achieve this high power efficiency by
maximizing the reuse of the values fetched from DRAM: Kernels are kept
in SRAM until the the entire layer has been processed. 
Activation values---typically from a patch across all the feature maps centered 
on a location in the image---are kept in SRAM until all the values are used to process the associated patch of output
feature maps. That way, each value fetched from DRAM is used hundreds or thousands of times.
This reuse amortizes the energy cost of the DRAM memory access 
so that it becomes similar to a local SRAM access.
Once the layer pair has been processed, the kernels are not needed again for that frame of input.
The result is that the arithmetic and memory access energy becomes comparable to that of SOA SNNs.
It makes sense, 
because they use the same transistors and logic circuits to do the operations.

In a synchronous CNN accelerator, 
once a layer has been used to compute all its downstream targets, it can be discarded.
It means that synchronous layer-by-layer CNN accelerators do not need to store 
the entire network activation state. A DRAM-based accelerator has the additional advantage that
it can share the memory and its interface
with other applications.

\section{Conclusion}
% Our experiences with NullHop and DeltaRNN were valuable because they showed
% that the data-driven sparsity principles of SNNs are equally valuable for
% synchronous DNN accelerators.

Starting from SNNs, we found out that the key principles 
of data-driven sparse computing can equally benefit 
more flexible synchronous DNN accelerators.
There is still potential 
in exploiting more data sparsity,
not only in activations but also in the network connections.
This area is very active 
% and is usually termed \textit{sparsity-aware}
in the mainstream
accelerator community.

%  Developers of asynchronous SNNs privately acknowledge the drawbacks of incompatibility, 
% inflexibility and poor area efficiency and naturally do not point them out in their publications, or to
% to their funding agencies or investors.
% SNN developers rightfully point out positive features
% like data-driven computing, good time to inference
% and energy per inference, and 
% focus their work on discovering new 
% computational principles for inference and learning that 
% might stem from precise timing and local learning rules.

The availability of new memory technology 
could bring memory closer to computation,
and even swap the roles: It is possible that memory itself will 
do  the processing rather than the way we have it now~\cite{ankit2019puma}.
But this step will require 
a huge change in system architectures 
and will probably require a compelling application for 
the required investment at all levels of the supply chain.
% The most likely mass production SNN would be a standalone ASIC
% that does a particular
% problem cheaper, faster, and better than anything else.

\subsection*{Acknowledgments}
Group members D. Neil, B. Rueckauer, A. Aimar, and C. Gao did the bulk of 
the work to implement the ideas presented here. 
We also thank our colleague A. Linares-Barranco at Univ. of Seville and his students
for enjoyable collaboration on FPGA demonstrations.
This work was partially funded by the Samsung Global Research \textit{Neuromorphic Processor Project}. 
Additional support was provided by the University of Zurich and ETH Zurich.

\ifCLASSOPTIONcaptionsoff
  \newpage
\fi

\bibliographystyle{IEEEtran}
\bibliography{main}

\end{document}